\documentclass[letterpaper, 10 pt, conference]{ieeeconf} 
\IEEEoverridecommandlockouts                              
\usepackage{xcolor}
\newcommand{\algoName}{LbW\xspace}
\usepackage{graphicx}
\usepackage{subcaption}
\usepackage{float}
\usepackage[font={small}]{caption}
\usepackage{lscape}                                         
\usepackage{flushend}

\usepackage[lined,ruled,linesnumbered]{algorithm2e}

\usepackage{bm}                          
\usepackage{mathtools}
\usepackage{amsmath, amssymb, amscd}
\usepackage{wasysym} 
\usepackage{nicefrac}       
\usepackage{mdwmath}
\usepackage{mdwtab}
\usepackage{eqparbox} 

\usepackage{array} 
\usepackage{tabularx}
\usepackage{multirow}
\usepackage{multicol}
\usepackage{booktabs}   
\usepackage{colortbl}

\usepackage{paralist}

\usepackage{epsfig}                      
\usepackage{graphicx}                  

\usepackage{color}

\usepackage{comment}
\usepackage{soul} 

\makeatletter
\let\NAT@parse\undefined
\makeatother

\usepackage{url}  
\usepackage[pagebackref=false,breaklinks=true,colorlinks,urlcolor=blue,citecolor=blue,linkcolor=blue,bookmarks=false]{hyperref}

\usepackage{listings}

\usepackage{xspace}
\usepackage{setspace}

\usepackage{nicefrac}
\usepackage{microtype}
\usepackage[utf8]{inputenc} 
\usepackage[T1]{fontenc}    

\usepackage[sort,compress]{cite}

\renewcommand{\baselinestretch}{1}       

\def\etal{et~al\mbox{.}\ }

\newlength\paramargin
\newlength\figmargin
\newlength\tablemargin
\newlength\secmargin
\newlength\figcapmargin
\newlength\rowmargin

\newlength\twoimg

\definecolor{napiergreen}{rgb}{0.16, 0.5, 0.0}

\newcommand{\heading}[1]
{\noindent\textbf{#1}}

\newcommand{\mpage}[2]
{
\begin{minipage}{#1\linewidth}\centering
#2
\end{minipage}
}

\definecolor{pink}{RGB}{219, 48, 122}

\long\def\ignorethis#1{}



\graphicspath{{figure}, {example}}

\title{\LARGE \bf 
Learning by Watching: Physical Imitation of \\ Manipulation Skills from Human Videos
}

\author{
Haoyu Xiong$^{*\dagger}$,
Quanzhou Li$^{*}$,
Yun-Chun Chen$^{*}$,
Homanga Bharadhwaj$^{*}$,
Samarth Sinha$^{*}$,
Animesh Garg$^{*\ddagger}$
\thanks{$^{*}$University of Toronto \& Vector Institute, $^{\dagger}$Tianjin University, $^{\ddagger}$Nvidia.}

}

\begin{document}

\maketitle
\thispagestyle{empty}
\pagestyle{empty}

\begin{abstract}
%
%
Learning from visual data opens the potential to accrue a large range of manipulation behaviors by leveraging human demonstrations without specifying each of them mathematically, but rather through natural task specification. 
%
%
In this paper, we present Learning by Watching (LbW), an algorithmic framework for policy learning through imitation from a single video specifying the task. 
The key insights of our method are two-fold.
First, since the human arms may not have the same morphology as robot arms, our framework learns unsupervised human to robot translation to overcome the morphology mismatch issue.
Second, to capture the details in salient regions that are crucial for learning state representations, our model performs unsupervised keypoint detection on the translated robot videos.
The detected keypoints form a structured representation that contains semantically meaningful information and can be used directly for computing reward and policy learning. 
We evaluate the effectiveness of our LbW framework on five robot manipulation tasks, including reaching, pushing, sliding, coffee making, and drawer closing. 
Extensive experimental evaluations demonstrate that our method performs favorably against the state-of-the-art approaches. 
More results and analysis are available at \href{https://www.pair.toronto.edu/lbw-kp/}{\texttt{pair.toronto.edu/lbw-kp/}}.
\end{abstract}

\section{Introduction}

Robotic \emph{Imitation Learning}, also known as \emph{Learning from Demonstration} (LfD), allows robots to acquire manipulation skills performed by expert demonstrations through learning algorithms~\cite{kinesthetic1,kinesthetic2}.
While progress has been made by existing methods, collecting expert demonstrations remains expensive and challenging as it assumes access to both observations and actions via kinesthetic teaching~\cite{kinesthetic1,kinesthetic2}, teleoperation~\cite{teleop1,teleop2}, or crowdsourcing platform~\cite{Mandlekar2018ROBOTURKAC, Mandlekar2020IRISIR, Mandlekar2020LearningTG, Mandlekar2020HumanintheLoopIL}.
In contrast, humans have the ability to imitate manipulation skills by \textit{watching} third-person performances.
Motivated by this, recent methods resort to endowing robots with the ability to learn manipulation skills via physical imitation from human videos~\cite{liu2018imitation,avid,NIPS2019_8528}.

Unlike conventional LfD methods~\cite{kinesthetic1,kinesthetic2, teleop1, teleop2}, which assume access to both expert obsevations and actions, approaches based on imitation from human videos relax the dependencies, requiring \emph{only} human videos as supervision~\cite{liu2018imitation,avid,NIPS2019_8528}.
One of the main challenges of these imitation learning methods is how to minimize the domain gap between humans and robots. 
For instance, human arms may have different morphologies than those of robot arms.
To overcome the morphology mismatch issue, existing imitation learning methods~\cite{liu2018imitation,avid,NIPS2019_8528} typically leverage image-to-image translation models (e.g., CycleGAN~\cite{cyclegan}) to translate videos from the human domain to the robot domain.
However, simply adopting vanilla image-to-image translation models still does not solve the imitation from human videos task, since the image-to-image translation models often capture only the macro features at the expense of neglecting the details in salient regions that are crucial for downstream tasks~\cite{bau2019seeing}.

\begin{figure}[t]
  \begin{center}
    \includegraphics[width=1.0\linewidth]{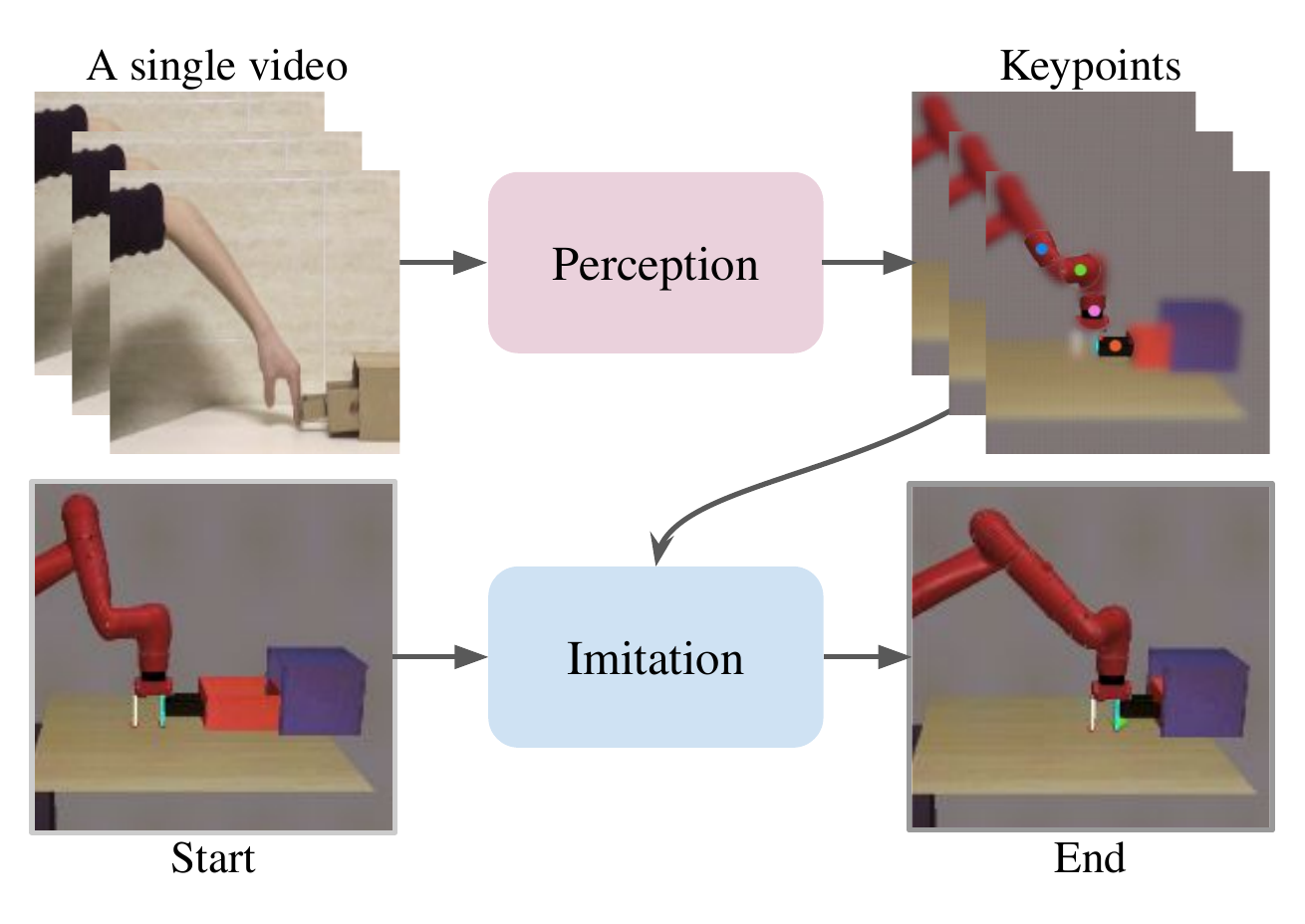} 
  \end{center}
  \vspace{-4.0mm}
  \caption{
  \textbf{LbW.}
  Given a single human video, our LbW framework learns human to robot translation followed by unsupervised keypoint detection.
  The resulting keypoint-based representations are semantically meaningful and can be used to guide the robot to learn manipulation skills through physical imitation.
  }
  \label{fig:teaser}
\end{figure}

In this paper, we present Learning by Watching (LbW), a framework for physical imitation from human videos for learning robot manipulation skills.
As shown in Figure~\ref{fig:teaser}, our framework is composed of a perception module and a policy learning module for physical imitation.
The perception module aims at minimizing the domain gap between the human domain and the robot domain as well as capturing the details of salient regions that are crucial for downstream tasks.
To achieve this, our perception module learns to translate the input human video to the robot domain with an unsupervised image-to-image translation model, followed by performing \emph{unsupervised} keypoint detection on the translated robot video.
The detected keypoints then serve as a structured representation that contains semantically meaningful information and can be used as input to the downstream policy learning module.

To learn manipulation skills, we cast this as a \emph{reinforcement learning} (RL) problem, where we aim to enable the robot to perform physically viable learning with the objective to imitate similar behavior as demonstrated in the translated robot video under context-specific constraints.
We evaluate the effectiveness of our LbW framework on five robot manipulation tasks, including reaching, pushing, sliding, coffee making, and drawer closing in two simulation environments (i.e., the Fetch-Robot manipulation in OpenAI gym~\cite{her} and meta-world~\cite{yu2019meta}).
Extensive experimental results show that our algorithm compares favorably against the state-of-the-art approaches.

The main contributions are summarized as follows:

\begin{enumerate}
  \item We present a framework for physical imitation from human videos for learning robot manipulation skills.
  \item Our method learns structured representations based on unsupervised keypoint detection that can be used directly for computing task reward and policy learning.
  \item Extensive experimental results show that our LbW framework achieves the state of the art on five robot manipulation tasks.
\end{enumerate}

\section{Related Work}


\heading{Imitation from human videos.}
Existing imitation learning approaches collect demonstrations by kinesthetic teaching~\cite{kinesthetic1,kinesthetic2}, teleoperation~\cite{teleop1,teleop2}, or through crowdsourcing platform~\cite{Mandlekar2018ROBOTURKAC, Mandlekar2020IRISIR, Mandlekar2020LearningTG, Mandlekar2020HumanintheLoopIL}, and assume access to both expert observations and expert actions at every time step. 
Recent progress in deep representation learning has accelerated the development of imitation from videos~\cite{NIPS2019_8528,avid, liu2018imitation,lin2020concept,tcn, pathakICLR18zeroshot, Unsup_percep_rew, meta_lfo_Yu, 2018-TOG-SFV, pmlr-v100-sieb20a, Sieb2018DataDF, schmeckpeper2020rlv, petrik2020real2sim}.
While applying image-to-image translation models to achieve imitation from human videos has been explored~\cite{liu2018imitation, NIPS2019_8528, schmeckpeper2020rlv}, the dependency on paired human-robot training data makes these methods hard to scale.

Among them, AVID~\cite{avid} is closely related to our work which translates human demonstrations to robot domain via CycleGAN~\cite{cyclegan} in an unpaired data setting. 
However, directly encoding the translated images using a feature extractor for deriving state representations may suffer from visual artifacts generated by image-to-image translation models, leading to suboptimal performance on downstream tasks.

Different from methods based on image-to-image translation models, Maximilian~\etal~\cite{pmlr-v100-sieb20a} leverage 3D detection to minimize the visual gap between the human domain and the robot domain. 
SFV~\cite{2018-TOG-SFV} enables humanoid characters to learn skills from videos based on deep pose estimation. 
Our method shares a similar reward computing scheme as these approaches~\cite{pmlr-v100-sieb20a, 2018-TOG-SFV, petrik2020real2sim, Sieb2018DataDF}.
The difference is that these methods require additional label data, whereas our framework is learned in an unsupervised fashion.

\heading{Cycle consistency.}
The idea of exploiting cycle consistency constraints has been widely applied in the context of image-to-image translation.
CycleGAN~\cite{cyclegan} learns to translate images in an unpaired data setting by exploiting the idea of cycle consistency.
UNIT~\cite{unit} achieves image-to-image translation by assuming a shared latent space between the two domains.
Other methods explore translating images across multiple domains~\cite{stargan} or learning to generate diverse outputs~\cite{bicycle,drit,munit}.
Recently, the idea of cycle consistency is also applied to address various problems such as domain adaptation~\cite{rao2020rl,rcan,crdoco,zhang2021learning} and policy learning~\cite{avid,gamrian2019transfer}. 
In our work, our LbW framework employs a MUNIT~\cite{munit} model to perform human to robot translation for achieving physical imitation from human videos.
We note that other unpaired image-to-image translation models are also applicable in our task.
We leave the discussion on the effect of different image-to-image translation models as future work.

\heading{Unsupervised keypoint detection.}
Detecting keypoints from images without supervision has been studied in the literature~\cite{zhang2018unsupervised,jakab2018unsupervised}.
In the context of computer vision, existing methods typically infer keypoints by assuming access to the temporal transformation between video frames~\cite{zhang2018unsupervised} or employing a differentiable keypoint bottleneck network without access to frame transition information~\cite{jakab2018unsupervised}.
Other approaches estimate keypoints based on the access to known image transformations and dense correspondences between features~\cite{thewlis2017unsupervised,shu2018deforming,wiles2018self}.

Apart from the aforementioned approaches, some recent methods focus on learning keypoint detection for image-based control tasks~\cite{transporter, kpintofuture, minderer2019unsupervised, finn2016deep}.
In our method, we adopt Transporter~\cite{transporter} to detect keypoints from the translated robot video in an unsupervised manner.
We note that while other unsupervised keypoint detection methods can also be used in our framework, the focus of our paper lies in learning structured representations that are semantically meaningful and can be used directly for downstream policy learning.
We leave the development of unsupervised keypoint detection methods as future work.

\section{Preliminaries} \label{sec:prelim}


To achieve physical imitation from human videos, we decompose the problem into a series of tasks: 1) human to robot translation, 2) unsupervised keypoint-based representation learning, and 3) physical imitation with RL. Here, we review the first two tasks, in which our method builds upon existing algorithms.

\begin{figure*}[t]
  \begin{center}
    \includegraphics[width=1.0\linewidth]{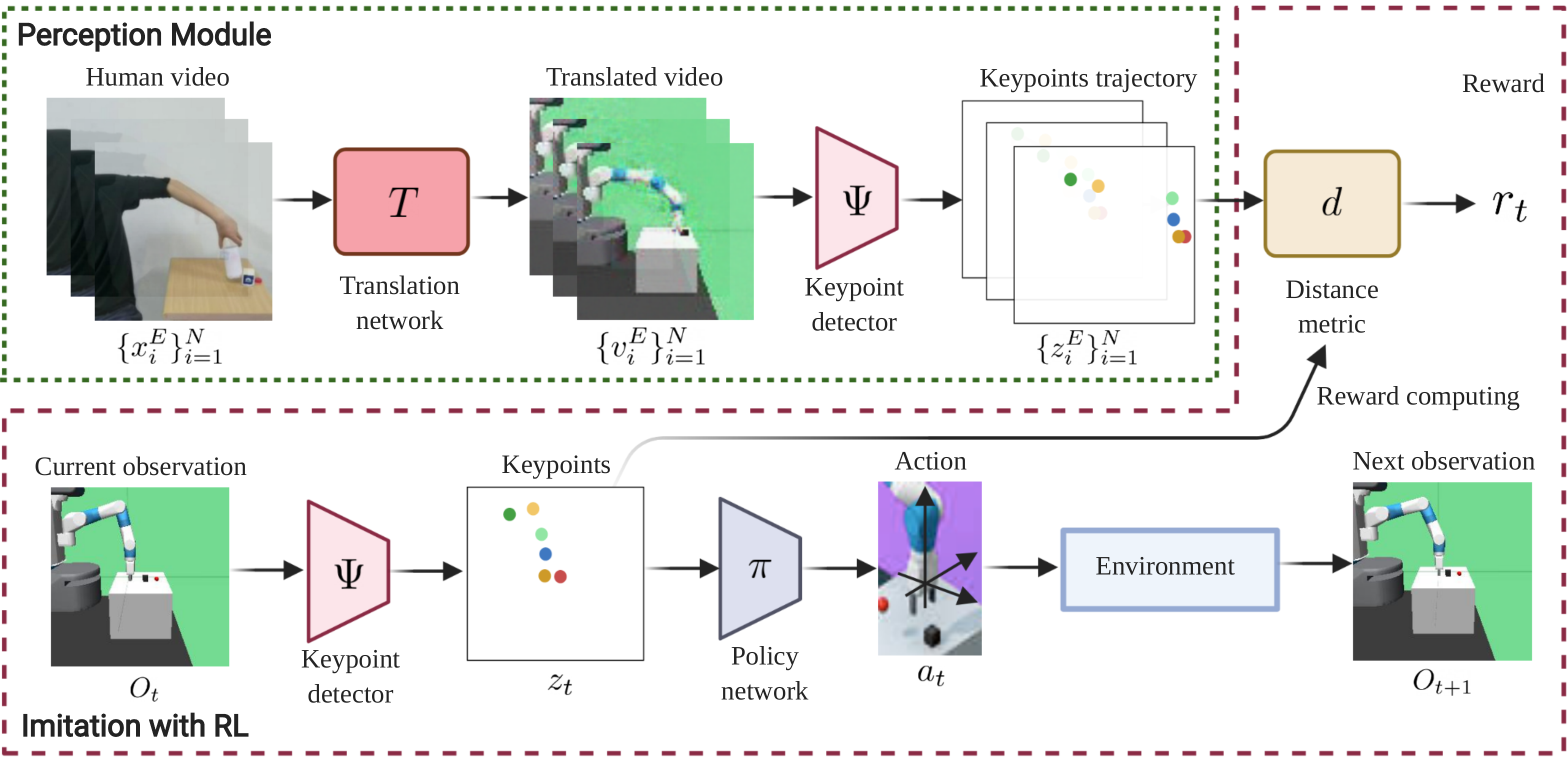}
  \end{center}
  \vspace{0.0mm}
  \caption{
  \textbf{Overview of the proposed LbW.} 
  Our LbW framework is composed of three main components: an image-to-image translation network $T$, a keypoint detector $\Psi$, and a policy network $\pi$.
  The image-to-image translation network translates the input human demonstration video frame by frame to generate a robot demonstration video.
  Next, the keypoint detector takes the generated robot demonstration video as input and extracts the keypoint-based representation for each frame to form a keypoints trajectory.
  At each time step, the keypoint detector also extracts the keypoint-based representation for the current observation.
  The reward for physical imitation is defined by a distance metric $d$ that measures the distance between the keypoint-based representation of the current observation and those in the keypoints trajectory.
  Finally, the keypoint-based representation of the current observation is passed to the policy network to predict an action that is used to interact with the environment.
  }
  \label{fig:method}
\end{figure*}

\subsection{Unsupervised Image-to-Image Translation} \label{sec:munit}

Similar to existing methods~\cite{avid,liu2018imitation}, we cast human to robot translation as an unsupervised image-to-image translation problem.
Specifically, we aim to learn a model that translates images from a source domain $X$ (e.g., human domain) to a target domain $Y$ (e.g., robot domain) \emph{without} paired training data~\cite{cyclegan,munit,drit,unit}.
In our method, we adopt MUNIT~\cite{munit} as the image-to-image translation network to achieve human to robot translation.
MUNIT learns to translate images between the two domains by assuming that an image representation can be disentangled into a domain-invariant content code (encoded by a content encoder $E^c$) and a domain-specific style code (encoded by a style encoder $E^s$).
The content encoders $E_X^c$ and $E_Y^c$ are shared in the two domains, whereas the style encoders $E_X^s$ and $E_Y^s$ of the two domains do \emph{not} share weights.
To translate an image from one domain to the other, we combine its content code with a style code sampled from the other domain.
The translations are learned to generate images that are indistinguishable from images in the translated domain.
Given an image $x$ from the source domain $X$ and an image $y$ from the target domain $Y$, we define the adversarial loss $\mathcal{L}_\mathrm{GAN}^x$ in the source domain as
\begin{equation}
  \mathcal{L}_\mathrm{GAN}^x = \mathbb{E}\bigg[\log D_X(x) + \log\Big(1 - D_X\big(G_X(c_y, s_x)\big)\Big)\bigg],
  \label{eq:adv-loss}
\end{equation}
where $c_y = E_Y^c(y)$ is the content code of image $y$, $s_x = E_X^s(x)$ is the style code of image $x$, $G_X$ is a generator that takes as input a content code $c_y$ and a style code $s_x$ and generates images that have similar distributions like those in the source domain, and $D_X$ is a discriminator that aims to distinguish between the translated images generated by $G_X$ and the images in the source domain.
The adversarial loss $\mathcal{L}_\mathrm{GAN}^y$ in the target domain can be similarly defined.

In addition to the adversarial losses, MUNIT applies reconstruction losses on images and content and style codes to regularize the model learning.
For the source domain, the image reconstruction loss $\mathcal{L}_\mathrm{rec}^x$ is defined as
\begin{equation}
  \mathcal{L}_\mathrm{rec}^x = \mathbb{E}\Big[\big\|G_X(c_x, s_x) - x\big\|\Big],
  \label{eq:rec-image-loss}
\end{equation}
the content reconstruction loss $\mathcal{L}_\mathrm{rec}^{c_x}$ is defined as
\begin{equation}
  \mathcal{L}_\mathrm{rec}^{c_x} = \mathbb{E}\Big[\big\|E_Y^c\big(G_Y(c_x, s_y)\big) - c_x\big\|\Big],
  \label{eq:rec-content-loss}
\end{equation}
and the style reconstruction loss $\mathcal{L}_\mathrm{rec}^{s_x}$ is defined as
\begin{equation}
  \mathcal{L}_\mathrm{rec}^{s_x} = \mathbb{E}\Big[\big\|E_X^s\big(G_X(c_y, s_x)\big) - s_x\big\|\Big].
  \label{eq:rec-style-loss}
\end{equation}
The image reconstruction loss $\mathcal{L}_\mathrm{rec}^y$, the content reconstruction loss $\mathcal{L}_\mathrm{rec}^{c_y}$, and the style reconstruction loss $\mathcal{L}_\mathrm{rec}^{s_y}$ in the target domain can be derived similarly.

The total loss $\mathcal{L}_\mathrm{MUNIT}$ for training MUNIT is given by
\begin{equation}
  \begin{split}
    \mathcal{L}_\mathrm{MUNIT} & = \mathcal{L}_\mathrm{GAN}^x + \mathcal{L}_\mathrm{GAN}^y + \lambda_\mathrm{image}(\mathcal{L}_\mathrm{rec}^x + \mathcal{L}_\mathrm{rec}^y) \\
    + & \lambda_\mathrm{content}(\mathcal{L}_\mathrm{rec}^{c_x} + \mathcal{L}_\mathrm{rec}^{c_y}) + \lambda_\mathrm{style}(\mathcal{L}_\mathrm{rec}^{s_x} + \mathcal{L}_\mathrm{rec}^{s_y}),
  \end{split}
  \label{eq:munit-loss}
\end{equation}
where $\lambda_\mathrm{image}$, $\lambda_\mathrm{content}$, and $\lambda_\mathrm{style}$ are hyperparameters used to control the relative importance of the respective loss functions.

\subsection{Unsupervised Keypoint Detection} \label{sec:transporter}

To perform control tasks, existing approaches typically resort to learning state representations based on image observations~\cite{lesort2018state,slac,solar,hafner2019learning,avid}.
However, the image observations generated by image-to-image translation models often capture only macro features while neglecting the details in salient regions that are crucial for downstream tasks.
Deriving state representations by encoding the translated image observations using a feature encoder would lead to suboptimal performance.
On the other hand, existing methods may also suffer from visual artifacts generated by the image-to-image translation models.
In contrast to these approaches, we leverage Transporter~\cite{transporter} to detect the keypoints in each translated video frame in an unsupervised fashion.
The detected keypoints form a structured representation that captures the robot arm pose and the location of the interacting object, providing semantically meaningful information for downstream control tasks while avoiding the negative impact of visual artifacts caused by the imperfect image-to-image translation.

To realize the learning of unsupervised keypoint detection, Transporter leverages object motion between a pair of video frames to transform a video frame into the other by transporting features at the detected keypoint locations.
Given two video frames $x$ and $y$, Transporter first extracts feature maps $\Phi(x)$ and $\Phi(y)$ for both video frames using a feature encoder $\Phi$ and detects $K$ $2$-dimensional keypoint locations $\Psi(x)$ and $\Psi(y)$ for both video frames using a keypoint detector $\Psi$.
Transporter then synthesizes the feature map $\hat{\Phi}(x, y)$ by suppressing the feature map of $x$ around each keypoint location in $\Psi(x)$ and $\Psi(y)$ and incorporating the feature map of $y$ around each keypoint location in $\Psi(y)$:
\begin{equation}
  \hat{\Phi}(x, y) = (1 - \mathcal{H}_{\Psi(x)}) \cdot (1 - \mathcal{H}_{\Psi(y)}) \cdot \Phi(x) + \mathcal{H}_{\Psi(y)} \cdot \Phi(y),
  \label{eq:transporting}
\end{equation}
where $\mathcal{H}_{\Psi(\cdot)}$ is a Gaussian heat map with peaks centered at each keypoint location in $\Psi(\cdot)$.

Next, the transported feature $\hat{\Phi}(x, y)$ is passed to a refinement network $R$ to reconstruct to the video frame $y$.
We define the loss $\mathcal{L}_\mathrm{transporter}$ for training Transporter as
\begin{equation}
  \mathcal{L}_\mathrm{transporter} = \mathbb{E}\Big[\big\|R\big(\hat{\Phi}(x, y)\big) - y\big\|\Big].
  \label{eq:transporter-loss}
\end{equation}
In the next section, we leverage the Transporter model to detect keypoints for each translated video frame.
The detected keypoints are then used as a structured representation for defining the reward function and as the input of the policy network to predict an action that is used to interact with the environment.

\section{Proposed Method} \label{sec:proposed method}

In this section, we first provide an overview of our approach.
We then describe the unsupervised domain transfer with keypoint-based representations module.
Finally, we describe the details of physical imitation with RL.

\subsection{Algorithmic Overview}

We consider the task of physical imitation from human videos for learning robot manipulation skills.
In this setting, we assume we have access to a \emph{single} human demonstration video $V_X = \{x_i^E\}_{i=1}^N$ of length $N$ depicting a human performing a specific task (e.g., pushing a block) that we want the robot to learn from, where $x_i^E \in \mathbb{R}^{H \times W \times 3}$ and $H \times W$ is the spatial size of $x_i^E$.
We note that the human actions are \emph{not} given in our setting.
Our goal is to develop a learning algorithm that allows the robot to \emph{imitate} the behavior demonstrated by the human in the human demonstration video $V_X$.
To achieve this, we present LbW, a framework that comprises three components: 1) the image-to-image translation network $T$ (from MUNIT~\cite{munit}), 2) the keypoint detector $\Psi$ (from the keypoint detector of Transporter~\cite{transporter}), and 3) the policy network $\pi$.

As shown in Figure~\ref{fig:method}, given a human demonstration video $V_X$ and the current observation $O_t \in \mathbb{R}^{H \times W \times 3}$ at time $t$, we first apply the image-to-image translation network $T$ to each frame $x_i^E$ in the human demonstration video $V_X$ and translate $x_i^E$ to a robot demonstration video frame $v_i^E \in \mathbb{R}^{H \times W \times 3}$.
Next, the keypoint detector $\Psi$ takes each translated robot video frame $v_i^E$ as input and extracts the keypoint-based representation $z_i^E = \Psi(v_i^E) \in \mathbb{R}^{K \times 2}$, where $K$ denotes the number of keypoints.
Similarly, we also apply the keypoint detector $\Psi$ to the current observation $O_t$ to extract the keypoint-based representation $z_t = \Psi(O_t) \in \mathbb{R}^{K \times 2}$.
To compute the reward for physical imitation, we define a distance metric $d$ that computes the distances between the keypoint-based representation $z_t$ of the current observation $O_t$ and each of the keypoint-based representations $z_i^E$ of the translated robot video frames $v_i^E$.
Finally, the policy network $\pi$ takes as input the keypoint-based representation $z_t$ of the current observation $O_t$ to predict an action $a_t = \pi(z_t)$ that is used to guide the robot to interact with the environment.
The details of each component are described in the following subsections.

\begin{figure}[t]
  \begin{center}
    \includegraphics[width=1.0\linewidth]{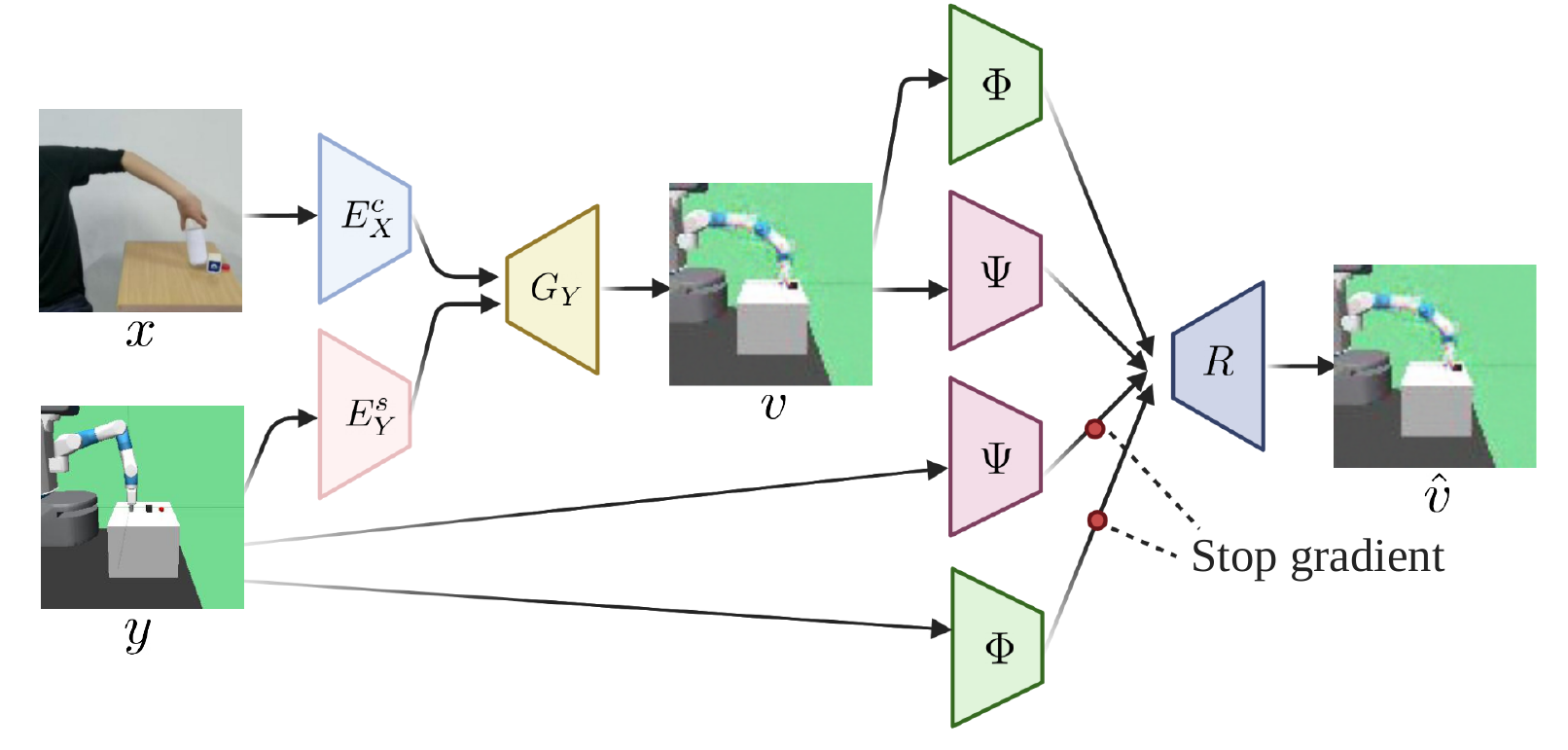}
  \end{center}
  \vspace{-2.0mm}
  \caption{
  \textbf{Overview of the perception module.}
  Our perception module is composed of a MUNIT network (left) and a Transporter model (right).
  Given a human video frame $x$ and a robot video frame $y$, the MUNIT model first extracts the content code of the human video frame and the style code of the robot video frame.
  The MUNIT model then generates the translated robot video frame $v$ by combining the extracted content code and style code.
  Next, the Transporter model extracts the features and detects the keypoints for both the translated robot video frame $v$ and the input robot video frame $y$ and reconstructs the translated robot video frame $\hat{v}$ by transporting features at the detected keypoint locations.
  Note that the input robot video frame $y$ is from a robot video generated by using a random policy.
  }
  \label{fig:perception}
\end{figure}

\subsection{Unsupervised Domain Transfer with Keypoints}

To achieve physical imitation from human videos, we develop a perception module that consists of a MUNIT model for human to robot translation and a Transporter network for keypoint detection as shown in Figure~\ref{fig:perception}.
To train the MUNIT model, we first collect the training data for the source domain (i.e., human domain) and the target domain (i.e., robot domain).
The source domain contains the human demonstration video $V_X$ that we want the robot to learn from.
To increase the diversity of the training data in the source domain for facilitating the MUNIT model training, we follow AVID~\cite{avid} and collect a few \emph{random} data by asking the human to randomly move the hands above the table \emph{without} performing the task.
As for the target domain training data, we collect a number of robot videos generated by having the robot perform a number of actions that are randomly sampled from the action space.
As such, the collection of the robot videos does \emph{not} require human expertise and effort.

Using the training data from both source and target domains, we are able to train the MUNIT model to achieve human to robot translation using the total loss $\mathcal{L}_\mathrm{MUNIT}$ in \eqref{eq:munit-loss} and following the protocol described in Section~\ref{sec:munit}.
After training the MUNIT model, we are able to translate the human demonstration video $V_X = \{x_i^E\}_{i=1}^N$ frame by frame to the robot demonstration video $\{v_i^E\}_{i=1}^N$ by combining the content code of each human demonstration video frame and a style code randomly sampled from the robot domain.

\begin{figure*}[t]
  \begin{center}
    \mpage{0.01}{\raisebox{0pt}{\rotatebox{90}{Human videos}}}
    \mpage{0.225}{\includegraphics[width=\linewidth]{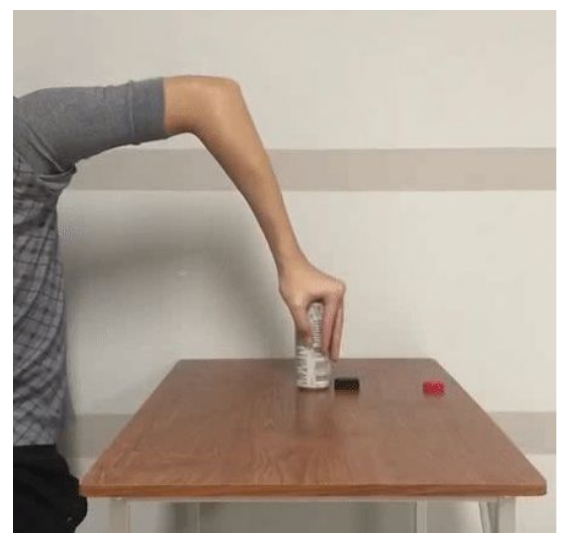}} \hfill
    \mpage{0.225}{\includegraphics[width=\linewidth]{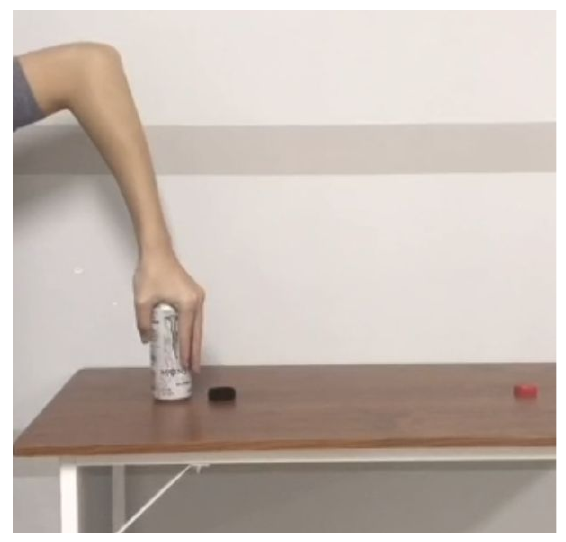}} \hfill
    \mpage{0.225}{\includegraphics[width=\linewidth]{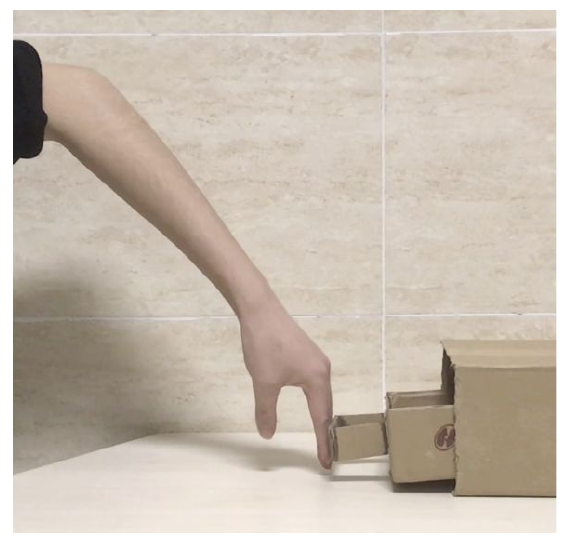}} \hfill
    \mpage{0.225}{\includegraphics[width=\linewidth]{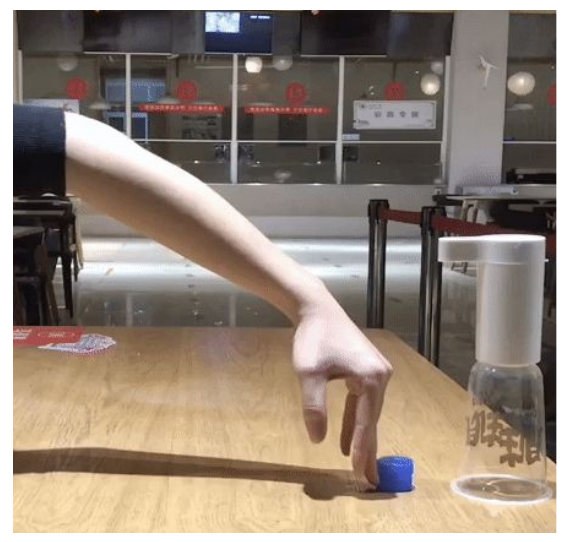}} \\
    \vspace{3.0mm}
    \mpage{0.01}{\raisebox{0pt}{\rotatebox{90}{Robot observations}}}
    \mpage{0.225}{\includegraphics[width=\linewidth]{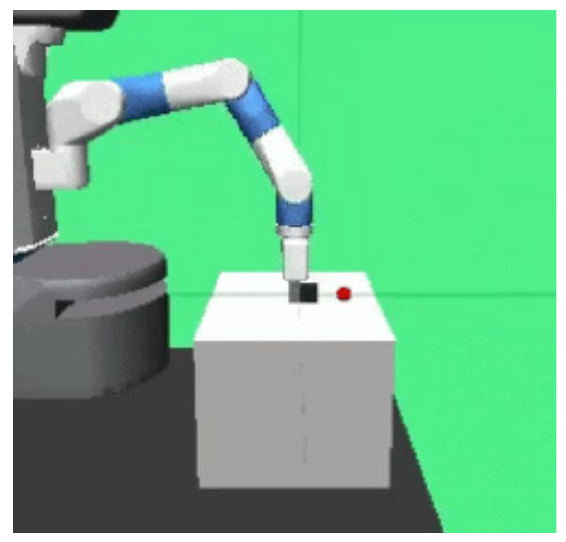}} \hfill
    \mpage{0.225}{\includegraphics[width=\linewidth]{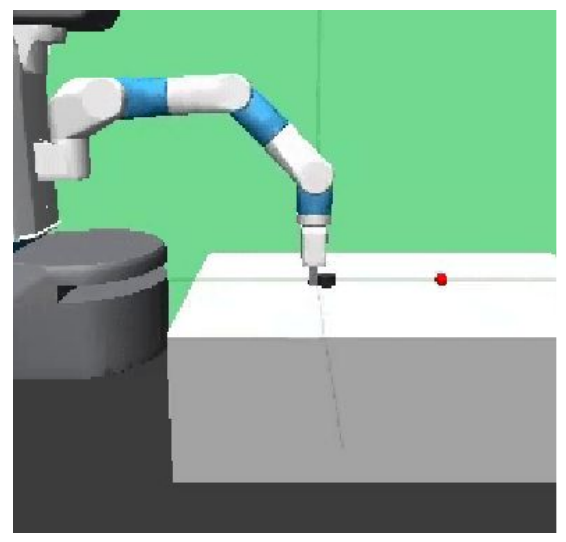}} \hfill
    \mpage{0.225}{\includegraphics[width=\linewidth]{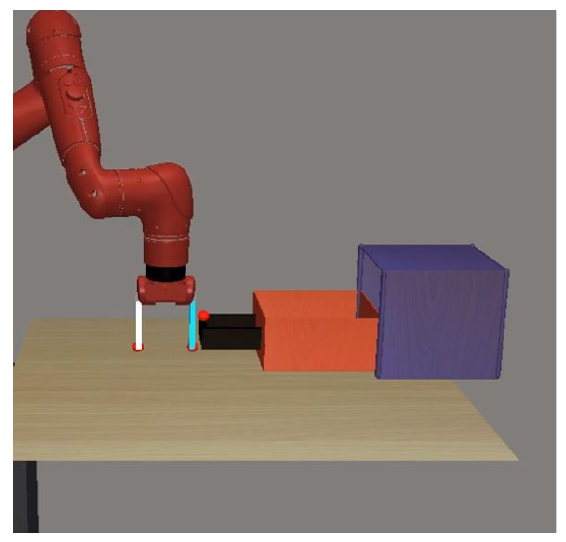}} \hfill
    \mpage{0.225}{\includegraphics[width=\linewidth]{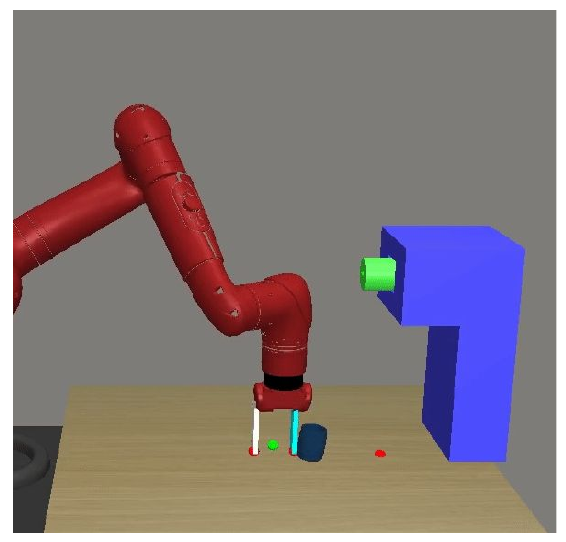}} \\
    \vspace{1.0mm}
    \mpage{0.01}{\raisebox{0pt}{}}
    \mpage{0.225}{Pushing} \hfill
    \mpage{0.225}{Sliding} \hfill
    \mpage{0.225}{Drawer closing} \hfill
    \mpage{0.225}{Coffee making} \\
    \vspace{1.0mm}
    \captionof{figure}{
    \textbf{Task overview.}
    We present the sample task scenes and one sample human video frame for the pushing, sliding, drawer closing, and coffee making tasks, respectively.
    Our human videos can be collected in an environment with a plain background (i.e., the left three columns) or with a noisy background (i.e., the rightmost column).
    }
    \label{fig:exp-tasks}
  \end{center}
\end{figure*}

As mentioned in Section~\ref{sec:transporter}, we aim to learn keypoint-based representations from the translated robot video in an unsupervised fashion.
To achieve this, we leverage Transporter to detect the keypoints in each translated robot video frame in an unsupervised fashion, as there are no ground-truth keypoint annotations available.

As illustrated in Figure~\ref{fig:perception}, following the protocol stated in Section~\ref{sec:transporter}, the Transporter model takes as input a translated robot demonstration video frame $v$ and a robot video frame $y$ from a robot video collected by applying a random policy and extracts their features and detects keypoint locations, respectively.
The Transporter model then reconstructs the translated robot demonstration video frame.
To train the Transporter model, we optimize the total loss $\mathcal{L}_\mathrm{transporter}$ in \eqref{eq:transporter-loss}.
Once the training of the Transporter model converges, we are able to use the keypoint detector $\Psi$ of the Transporter model to extract a keypoint-based representation $z_i^E = \Psi(v_i^E)$ for each frame $v_i^E$ in the translated robot demonstration video to form a keypoints trajectory $\{z_i^E\}_{i=1}^N$ and a keypoint-based representation $z_t = \Psi(O_t)$ for the current observation $O_t$.
The keypoints trajectory $\{z_i^E\}_{i=1}^N$ of the translated robot demonstration video $\{v_i^E\}_{i=1}^N$ and the keypoint-based representation $z_t$ of the current observation $O_t$ provide \emph{semantically} meaningful information for robot manipulation tasks.
We then use both of them to compute the reward $r_t$ and use the keypoint-based representation $z_t$ of the current observation $O_t$ to predict an action $a_t$.
The details of reward computing and policy learning are elaborated in the next subsection.

\subsection{Physical Imitation with RL}

To control the robot, we use RL to learn a policy from image-based observations that maximize the cumulative values of a learned reward function. 
In our method, we \emph{decouple} the policy learning phase from the keypoint-based representation learning phase.
Given the keypoints trajectory $\{z_i^E\}_{i=1}^N$ of the translated robot demonstration video $\{v_i^E\}_{i=1}^N$ and the keypoint-based representation $z_t$ of the current observation $O_t$, our policy network $\pi$ outputs an action $a_t = \pi(z_t)$ which is executed in the environment to obtain the next observation $O_{t+1}$. 
To achieve physical imitation, we aim to minimize the distance between the keypoints trajectory of the agent and that of the translated robot demonstration video.
Specifically, we define the reward $r_t$ as
\begin{equation}
  r_t = d\big(z_t, z_{t+1}, \{z_i^E\}_{i=1}^N\big) = \lambda_{r_1} \cdot r_1(t) + \lambda_{r_2} \cdot r_2(t),
  \label{eq:reward}
\end{equation}
where $\lambda_{r_1}$ and $\lambda_{r_2}$ are hyperparameters that balance the importance between the two terms, and the aforementioned goal is imposed on $r_1(t)$ and $r_2(t)$, which are defined by the following equations:
\begin{equation}
  r_1(t) = - \min \|z_t - z_p^E\|, \textmd{ and}
  \label{eq:reward_1}
\end{equation}
\begin{equation}
  r_2(t) = - \min\Big(\big\|(z_{t+1} - z_t) - (z_{q+1}^E - z_q^E)\big\|\Big),
  \label{eq:reward_2}
\end{equation}
where $1 \leq p \leq N-1$, $1 \leq q \leq N-1$, $r_1(t)$ aims to minimize the distance between the keypoint-based representation $z_t$ of the current observation $O_t$ and the most similar (closest) keypoint-based representation $z_p^E$ in the keypoints trajectory $\{z_i^E\}_{i=1}^N$ of the translated robot demonstration video $\{v_i^E\}_{i=1}^N$, and $r_2(t)$ is the first-order difference equation of $r_1(t)$.

We add the tuple $(z_t, a_t, z_{t+1}, r_t)$ to a replay buffer.
%
Then, the policy network $\pi$ can be trained with any RL algorithms in principle. We make use of Soft-Actor Critic (SAC)~\cite{sac} as the RL algorithm for policy learning in our experiments.



\begin{figure*}
  \begin{center}
    \mpage{0.01}{\raisebox{0pt}{\rotatebox{90}{Input}}}
    \mpage{0.17}{\includegraphics[width=\linewidth]{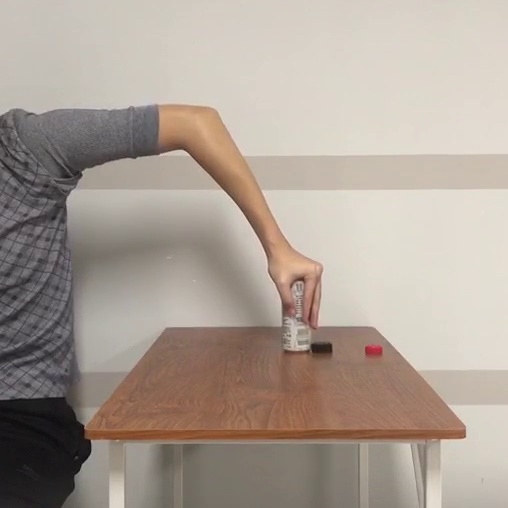}} \hfill
    \mpage{0.17}{\includegraphics[width=\linewidth]{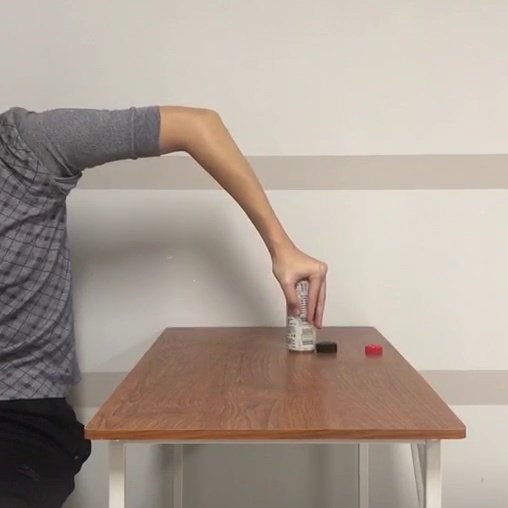}} \hfill
    \mpage{0.17}{\includegraphics[width=\linewidth]{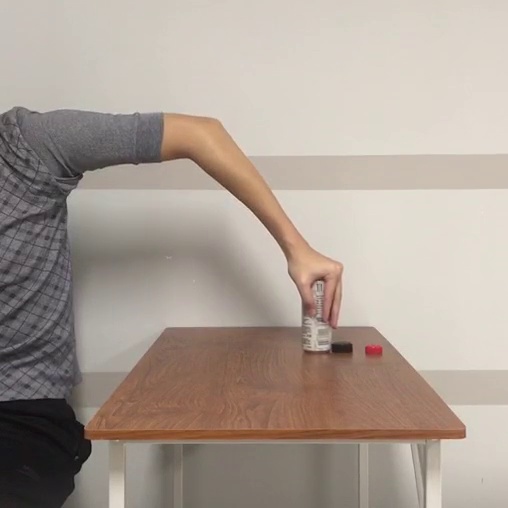}} \hfill
    \mpage{0.17}{\includegraphics[width=\linewidth]{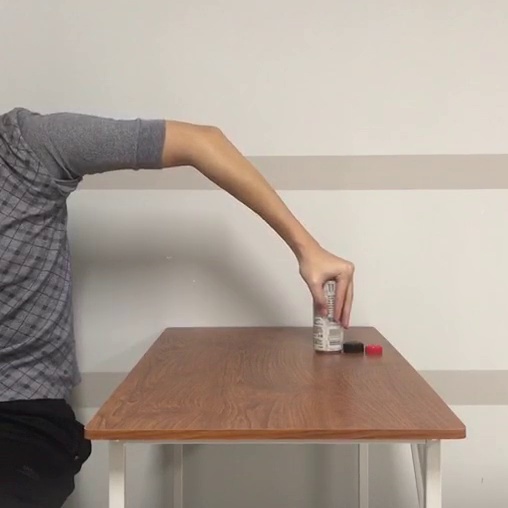}} \hfill
    \mpage{0.17}{\includegraphics[width=\linewidth]{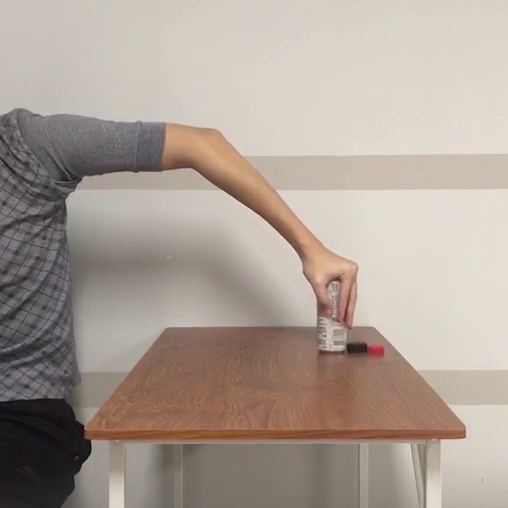}} \\
    \vspace{3.0mm}
    \mpage{0.01}{\raisebox{0pt}{\rotatebox{90}{CycleGAN}}}
    \mpage{0.17}{\includegraphics[width=\linewidth]{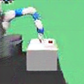}} \hfill
    \mpage{0.17}{\includegraphics[width=\linewidth]{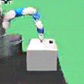}} \hfill
    \mpage{0.17}{\includegraphics[width=\linewidth]{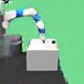}} \hfill
    \mpage{0.17}{\includegraphics[width=\linewidth]{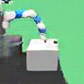}} \hfill
    \mpage{0.17}{\includegraphics[width=\linewidth]{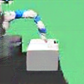}} \\
    \vspace{3.0mm}
    \mpage{0.01}{\raisebox{0pt}{\rotatebox{90}{LbW (Ours)}}} 
    \mpage{0.17}{\includegraphics[width=\linewidth]{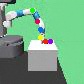}} \hfill
    \mpage{0.17}{\includegraphics[width=\linewidth]{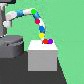}} \hfill
    \mpage{0.17}{\includegraphics[width=\linewidth]{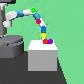}} \hfill
    \mpage{0.17}{\includegraphics[width=\linewidth]{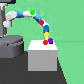}} \hfill
    \mpage{0.17}{\includegraphics[width=\linewidth]{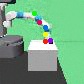}} \\
    \vspace{2.0mm}
    \captionof{figure}{
    \textbf{Visual results and comparisons on the pushing task.} 
    Given a human video as input in the first row, we present the translated images of CycleGAN~\cite{cyclegan} in the second row.
    In the third row, we visualize our translated images and the detected keypoints produced by the perception module.
    Our perception module accurately detects the robot arm pose and the location of the interacting object.
    }
    \label{fig:exp-perception}
  \end{center}
\end{figure*}

\section{Experiments}

In this section, we describe the experimental settings and report results with comparisons to state-of-the-art methods on five robot manipulation tasks.
Through experiments, we aim to investigate the following questions:
\begin{enumerate}
  \item How accurate is our perception module in handling the human-robot domain gap and in detecting keypoints?
  \item How does \algoName compare with state-of-the-art baselines in terms of performance on robot manipulation tasks?
\end{enumerate}
%

\subsection{Experimental Setting}


%
We perform experimental evaluations in two simulation environments, i.e., the Fetch-Robot manipulation in OpenAI gym~\cite{her} and meta-world~\cite{yu2019meta}.
We evaluate on five tasks: reaching, pushing, sliding, coffee making, and drawer closing.
Figure~\ref{fig:exp-tasks} presents the overview of each task, including the task scenes and one sample human video frame for each task.
The goal of each task is described as follows.

\begin{enumerate}
  \item For the \textit{reaching} task, the robot has to move its end-effector to reach the target.
  \item For the \textit{pushing} task, a puck is placed on the table in front of the robot, and the goal is to move the puck to the target location.
  \item For the \textit{sliding} task, a puck is placed on a long slippery table and the target location is beyond the reach of the robot. The goal is to apply an appropriate force to the puck so that the puck slides and stops at the target location due to friction.
  \item For the \textit{coffee making} task, a cup is placed on the table in front of the robot, and the goal is to move the cup to the location right below the coffee machine outlet. The moving distance of coffee making task is longer than the one in the pushing task.
  \item For the \textit{drawer closing} task, the robot has to move its end-effector to close the drawer. 
\end{enumerate}

In the policy learning phase, the robot receives only an RGB image of size $84 \times 84 \times 3$ as the observation. 
The robot arm is controlled by an Operational Space Controller in end-effector positions.
As each of the tasks is described by a single human video,
we set the initial locations of the object and the target to a fixed configuration.

\begin{table}[t]
  \begin{center}
  \normalsize
  \caption{
  \textbf{Dataset statistics for perception module training for each task.}
  We summarize the number of video frames of both the source (human) domain and the target (robot) domain for training the perception module for each task.
  }
  \label{exp:statistics}
  \resizebox{\linewidth}{!} 
  {
  \begin{tabular}{l|ccccc}
    \toprule
    Domain & Reaching & Pushing & Sliding & Drawer closing & Coffee making \\
    \midrule
    Source (human) & 1,056 & 398 & 650 & 986 & 658 \\
    Target (robot) & 3,150 & 1,220 & 2,120 & 2,940 & 4,007 \\
    \bottomrule
  \end{tabular}
  }
  \end{center}
\end{table}

\begin{table*}[t]
  \begin{center}
  \normalsize
  \caption{\textbf{Success rates.} Comparison of success rates for test evaluations of our LbW framework and the baselines.}
  \label{exp:results}
  \resizebox{\linewidth}{!} 
  {
  \begin{tabular}{l|c|ccccc}
    \toprule
    Method & Number of expert demonstrations & Reaching & Pushing & Sliding & Drawer closing & Coffee making \\
    \midrule
    Classifier reward & 35 robot videos & \textbf{100}\% & \textbf{100}\% & 30\% & 70\% & 50\% \\
    AVID-m & 15 human videos & \textbf{100}\% & 60\% & 0\% & 50\% & 40\% \\    
    LbW (Ours) & \textbf{1} human video & \textbf{100}\% & \textbf{100}\% & \textbf{80}\% & \textbf{80}\% & \textbf{70}\% \\
    \bottomrule
  \end{tabular}
  }
  \end{center}
\end{table*}

\subsection{Comparison to Baseline Methods}

To evaluate the effectiveness of our perception module, we implement two baseline methods using the same control model as LbW, which is adopted from SAC$+$AE~\cite{sac_ae}, but with different reward learning methods. 

\heading{Classifier-reward.}
We implement a classifier-based reward learning method in a similar way as VICE~\cite{vice}.
For each task, given robot demonstration videos, instead of the human videos, the CNN classifier is pre-trained on \emph{ground-truth} goal images with \emph{positive labels} and the remaining images with \emph{negative labels}.
To learn a policy in the environment, we adopt the implementation from SAC$+$AE\cite{sac_ae}, where we use the classifier-based reward to train the agent.

\heading{AVID-m.} 
Since AVID~\cite{avid} is the state-of-the-art method that outperforms prior approaches, including BCO~\cite{bco} and TCN~\cite{tcn}, we focus on comparing our method with AVID.
For a fair comparison, we reproduce the reward learning method of AVID and replace the control module with SAC$+$AE\cite{sac_ae}.
We denote this method as AVID-m. 
For each task, given human demonstration videos, we first translate the human demonstration videos to the robot domain using the CycleGAN~\cite{cyclegan} model. 
Then the CNN classifier is pre-trained on the \emph{translated} goal images with \emph{positive labels} and the remaining translated images with \emph{negative labels}.
For RL training, we adopt the implementation from SAC$+$AE~\cite{sac_ae}.

\subsection{Dataset Collection and Statistics}

We decouple the training phase of the perception module from that of the policy learning module.

\heading{Dataset for perception module training.} 
To train our perception module and the CycleGAN method, we collect human expert videos and videos of a human performing random actions without performing the tasks for the human domain.
For the robot domain, we first constrain the action space of the robot such that unexpected robot poses will not occur (i.e., robot arms are constrained to move above the table), and then run a random policy to collect robot videos. 
Note that we do \emph{not} use robot expert videos for training the perception module.
Table~\ref{exp:statistics} presents the dataset statistics for each task for training the perception module.

\heading{Dataset for policy learning.} 
For policy learning, we use only one single human expert video to train our policy network.
The AVID-m method uses $15$ human expert videos, while the classifier-reward approach uses $35$ robot expert videos.

%
%
%

\subsection{Performance Evaluations}

Following AVID~\cite{avid}, we use \emph{success rate} as the evaluation metric.
At test time, the task is considered to be a success if the robot is able to complete the task within a specified number of time steps (i.e., $50$ time steps for reaching and pushing, and $300$ time steps for sliding, coffee making, and drawer closing).
The results are evaluated by $10$ test episodes for each task. 
Table~\ref{exp:results} reports the success rates of our method and the two baseline approaches on all five tasks.
We find that for the reaching task, all three methods achieve a success rate of $100\%$.
For the sliding, drawer closing, and coffee making tasks, our LbW performs favorably against the two competing approaches.

The difference between the AVID-m method and the classifier-reward approach is that AVID-m leverages CycleGAN for human to robot translation, while the classifier-reward method using ground-truth robot images directly. 
As shown in Figure~\ref{fig:exp-perception}, the translated images of AVID-m have clear visual artifacts.
For instance, the red cube disappears and the robot poses in the translated images do not match those in the human video frames.
The comparisons between AVID-m and the classifier-reward method and the visual results of AVID-m in Figure~\ref{fig:exp-perception} show that using image-to-image translation models alone for minimizing the human-robot domain gap will have negative impact to the performance of the downstream tasks.
Our perception module learns unsupervised human to robot translation as well as unsupervised keypoint detection on the translated robot videos.
The learned keypoint-based representation provides semantically meaningful information for the robot, allowing our LbW framework compares favorably two competing approaches.
More results, videos, performance comparisons, and implementation details are available at \href{https://www.pair.toronto.edu/lbw-kp/}{\texttt{pair.toronto.edu/lbw-kp/}}.%

%
%
%

%
%
%

\subsection{Discussion of Limitations}

While results on five tasks demonstrate the effectiveness of our LbW framework, there are two limitations.
First, existing imitation learning methods~\cite{avid,liu2018imitation, NIPS2019_8528, schmeckpeper2020rlv} that are based on image-to-image translation require the pose of the human arms and that of the robot arms to be similar.
As a result, these methods may not perform well on human demonstration videos that have larger pose variations or with more natural poses.
Our LbW framework also leverages an image-to-image translation model, thus suffering from the same limitation as these methods.
Second, in our method, learning from only a \emph{single} human video limits the model generalization to new scenes.


\section{Conclusions}

We introduced LbW, a framework for physical imitation from human videos.
Our core technical novelty lies in the design of the perception module that minimizes the domain gap between the human domain and the robot domain followed by keypoint detection on the translated robot video frames in an unsupervised manner.
The resulting keypoint-based representations capture semantically meaningful information that guide the robot to learn manipulation skills through physical imitation.
We defined a reward function with a distance metric that encourages the trajectory of the agent to be as close to that of the translated robot demonstration video as possible.
Extensive experimental results on five robot manipulation tasks demonstrate the effectiveness of our approach and the advantage of learning keypoint-based representations over conventional state representation learning approaches.

\heading{Acknowledgement.}
Animesh Garg is supported by CIFAR AI chair, and we would like to acknowledge Vector institute for computation support. 

\bibliographystyle{ieeetr}
\bibliography{reference.bib}

\end{document}